\documentclass[letterpaper, 10 pt, conference]{ieeeconf}  

\IEEEoverridecommandlockouts                              

\usepackage{amssymb} 

\usepackage[table]{xcolor}

\usepackage{array}

\usepackage{placeins}

\usepackage{subcaption} 

\usepackage{multirow}

\usepackage{bm}

\usepackage{standalone}
\usepackage{tikz}
\usetikzlibrary{shapes.geometric, arrows.meta, positioning, calc, fit, backgrounds, decorations.pathreplacing}

\usepackage{hyperref}

\usepackage{mathtools}

\title{\LARGE \bf
Policy-Calibrated DAgger: Offline Calibrated Noise Injection for Imitation Learning
}

\author{Jenny Wang$^{1}$ and George Kantor$^{1}$
\thanks{$^{1}$J. Wang and G. Kantor are with Carnegie Mellon University, Pittsburgh PA, USA \{jennyw2, gkantor\}@andrew.cmu.edu}%
}

\begin{document}

\maketitle
\thispagestyle{empty}
\pagestyle{empty}

\begin{abstract}

Policies trained with imitation learning can accumulate errors over time, causing the robot to drift outside the training distribution. Existing methods mitigate this covariate shift by collecting additional data where the policy fails or is likely to fail. The first places the robot in unsafe conditions and the second requires choosing an appropriate noise distribution to collect new expert demonstrations under that noise. We propose Policy-Calibrated DAgger, a method that makes use of the properties of recent generative policies to estimate the policy's noise offline by using its own predicted action distribution. We measure a diffusion policy's spread of predicted actions at observations along the expert trajectory and measure its closed-loop error relative to a recorded trajectory. To address issues with measuring error in a multimodal action space, we guide the policy towards the trajectory during closed-loop control through partial denoising, and use properties of a diffusion model to unnormalize the measured error as if we did not guide it. We experiment in a scenario where a robot is tasked to reach an engine lever in a cluttered and narrow environment and show results in a 3D photorealistic simulator and a 2D planar reacher environment. We show that our method surpasses policies trained with dataset aggregation without noising and matches the performance of the best noise level in hindsight, without requiring a sweep over noise levels.

\end{abstract}

\section{Introduction}

Behavior cloning is a popular method that enables scalable learning for robot manipulation~\cite{chi2024universal, xiong2025vision, Khazatsky2024DROIDAL, Tao2025DexWildDH}.
However, its performance is limited by compounding errors over time, as small policy errors cause the state distribution to drift away from the training data~\cite{ross2011reduction, laskey2017dart}. DAgger mitigates this covariate shift by aggregating expert data within the policy's induced state distribution, yielding a policy whose errors scale linearly rather than quadratically with time~\cite{ross2011reduction}. DAgger relies on alternating control between the expert and the policy in order for the expert to provide guidance within the policy's induced state distribution. However, this makes the system unintuitive for a human expert to control. 

In practice, many methods do not set the frequency at which the expert is queried, but instead let the expert return control to the policy at their convenience~\cite{kelly2019hg, wu2025robocopilot, lee2025diff}. This enables higher quality expert data at the cost of covering states less aligned with the policy's induced state distribution. Recent success has been found by equipping human experts with teleoperation devices that track the real robot with force feedback, enabling human experts to give corrective actions closer to the state distribution induced by the policy~\cite{liu2025factr, bazhenov2025echo}. However, these methods place the burden of distribution matching on the expert, which is labor-intensive.


\begin{figure}
    \centering
    \includegraphics[width=0.90\linewidth]{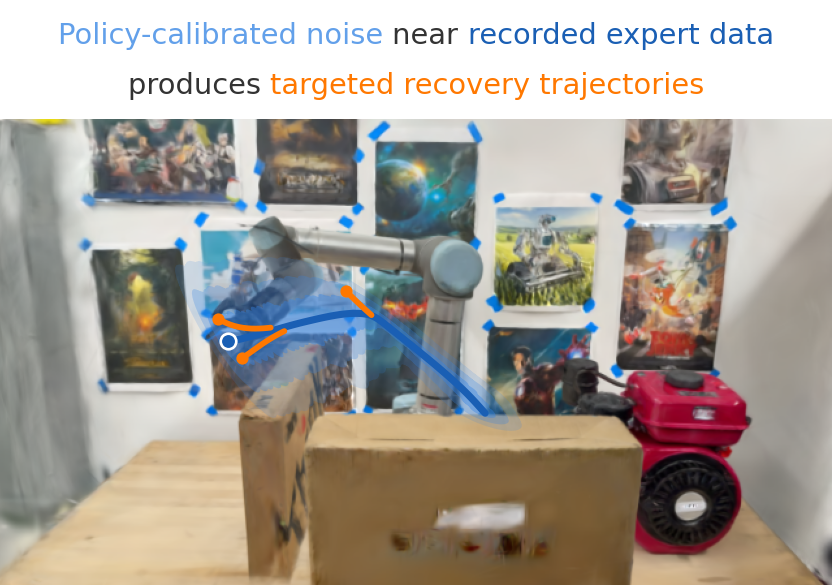}
    \caption{Our method estimates policy error while tracking a trajectory by making use of the probabilistic properties of generative policies. By sampling along \textcolor{blue}{pre-recorded expert data} and making use of the model's noise schedule to undo the normalizing effects of partial denoising towards the expert distribution in closed-loop control in simulation, we can estimate the policy's noise with respect to the expert data. We can sample \textcolor{orange}{targeted recovery trajectories} within this \textcolor{blue!60}{policy-calibrated noise} distribution to reduce covariate shift between the policy's visited states and the expert data.}
    \label{fig:overall-method-pic}
    \vspace{-0.3cm}
\end{figure}

Our key insight is that closing the gap between the state and action distributions induced by the expert and the policy is essential for addressing covariate shift in practice. This is important for long-horizon tasks requiring precise, full-body maneuvers like a robot arm reaching around a narrow and cluttered engine room. 
We propose a method that uses the policy to adapt expert intervention data to the policy's learned action distribution, without requiring the expert to control a noisy system or modifying policy training.
We build on partial diffusion~\cite{yoneda2023noise}, which adds noise to a sample of the output distribution and uses the model to denoise it to a similar result. We use it to simulate policy-noise-expert interactions probabilistically, without requiring the expert to adapt to a noisy system in the loop. 
Our policy-calibrated noise, when paired with Gaussian noise and action relabeling to connect augmented states back to the expert intervention, trains policies with better recovery behaviors by limiting covariate shift.

We draw close comparisons to data augmentation techniques for imitation learning that expand a limited set of expert demonstrations to cover the robot's workspace. Covering all possible states requires compute and modeling capacity. In addition, many approaches require an oracle policy or additional inductive biases about the environment. We show that our method expands the diversity of limited expert demonstrations in a way that targets task failures without needing an oracle policy or additional real-world data collection.

In this work, we propose a method to estimate policy-calibrated noise that reduces covariate shift between the state distributions induced by the policy and the expert. Our core contributions are as follows:

\begin{itemize}
    \item We propose policy-calibrated data augmentations, a data augmentation method that noises expert interventions to reduce covariate shift with respect to the policy's induced state distribution, generating highly relevant recovery state-action pairs.
    \item We extend partial denoising~\cite{yoneda2023noise} to action chunking.
    \item We show that our method increases data efficiency and policy performance in a long-horizon, cluttered goal-reaching task. Our experiments are performed on a photorealistic simulator of our real engine lever-reaching task mockup and also in a 2D planar simulator that encapsulates the narrow passageways in the real task.
\end{itemize}

\section{Related Work}

\subsection{Interactive Imitation Learning}
DAgger~\cite{ross2011reduction} methods address the data mismatch and compounding error issues in behavior cloning. In theory, DAgger at convergence will collect expert recovery actions in parts of the state space that the policy would visit. DART~\cite{laskey2017dart} shows that aggregating expert demonstrations within a certain noise level of the optimal policy reduces covariate shift. This theoretically mitigates the data mismatch problem. In practice, the constant switching between policy actions and expert actions makes it difficult for experts to provide corrective input due to system oscillations~\cite{Ross2012LearningMR}. Some methods tackle this data-intensive process with end-to-end automated world models~\cite{yu2026wmdagger, zhang2024diffusion}, only querying the expert when the model is uncertain~\cite{haridas2023dadagger, He2025UncertaintyCF, Hoque2021ThriftyDAggerBN, Lee2024DiffDaggerUE}, or when the state has high variance~\cite{Singi2023DecisionMF}. HG-DAgger~\cite{kelly2019hg} is a variant of DAgger that lets the expert decide whether to execute policy actions or their own actions in the system. The expert is instructed to give corrective inputs whenever the system enters an unsafe state, then return control to the policy once the system is safe again. Dataset aggregation methods designed with human-gated intervention tend to show strong empirical results~\cite{wu2025robocopilot, liu2025factr, bazhenov2025echo} despite long stretches of expert actions that may move the system far from the policy distribution. Recent work has focused on residual interventions and providing the human expert with force feedback to collect error intervention data without interrupting the current robot policy~\cite{xu2026compliant, wu2025robocopilot}. This allows the expert interventions to stay close to the policy's induced state distribution by design. However, this places the burden of matching the policy on the expert. We propose a method that makes the policy adapt to the expert.


\subsection{Learning from Suboptimal Data}

Some imitation learning methods treat suboptimal data as a noisy ground truth signal. For example, pretraining a model on an auxiliary task often enables data-efficient training on the target task if the two tasks are aligned~\cite{li2024visual, chen2024don, tao2025incremental}. Also, simulated data can supplement limited real-world training datasets to increase task success rate when co-training~\cite{maddukuri2025sim}. Iteratively fine-tuning a policy on a continuously changing robot embodiment can increase the efficacy of policy transfer from a data source with a large embodiment gap~\cite{liu2022revolver, liu2022herd}.

Other methods use noisy data to shape what expert intervention data should be collected. This can be done by querying the expert when a policy is uncertain~\cite{haridas2023dadagger, He2025UncertaintyCF, Hoque2021ThriftyDAggerBN, Lee2024DiffDaggerUE} or by injecting noise into the expert data distribution~\cite{laskey2017dart, ke2021grasping, zhou2023nerf}. We observe that the policy is a useful model of its own errors. We use partial diffusion to sample the noise model of the policy. Partial diffusion has previously been studied in shared autonomy~\cite{yoneda2023noise, sun2025flashback} and image editing~\cite{meng2021sdedit}. Its result is bounded by some distance to the original action~\cite{meng2021sdedit}. We show that this bound lets us estimate the policy's error relative to the expert while constraining the policy to move similarly to the expert during trajectory sampling in offline simulations.
In addition, we show that the targeted data diversity obtained by noise calibration can be used to improve the policy's robustness with local action augmentations without additional data collection.

\begin{figure*}
    \centering
    \includegraphics[width=0.92\linewidth]{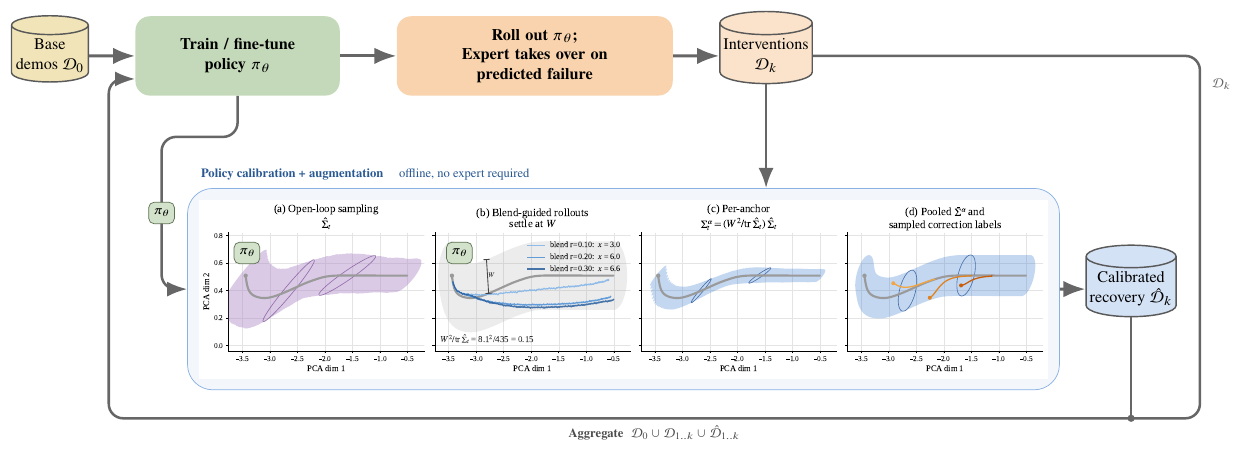}
    \vspace{-0.4cm}
    \caption{Our method augments the data collected during a DAgger-style data collection loop. The policy at data collection round $k$ is trained on the base dataset $\mathcal{D}_0$, aggregated intervention data $\mathcal{D}_1\ldots\mathcal{D}_k$ and noised intervention data $\hat{\mathcal{D}}_1\ldots\hat{\mathcal{D}}_k$. To generate policy-calibrated noised intervention data: \textbf{(a)} We sample the policy open loop at every other timestep along the expert trajectory to estimate $\hat\Sigma_t$. \textbf{(b)} We sample blended trajectories between the policy and expert with partial denoising to estimate $W$. \textbf{(c)} We combine (a) and (b) to compute the policy noise per timestep $\Sigma^{\alpha}_t$. \textbf{(d)} We average the per-timestep $\Sigma^{\alpha}_t$ across all data to obtain the pooled $\bar\Sigma^{\alpha}$. In this example, the per-timestep ellipsoidal noise estimates average to become more isotropic. We sample states and corrective action pairs within this local area around the expert intervention.}
    \label{fig:method_overview}
\vspace{-0.4cm}
    
\end{figure*}

\subsection{Data Augmentation for Imitation Learning}

Collecting demonstrations for imitation learning is a labor-intensive process. Many works propose methods to expand expensive expert data into a wider range of data through data augmentations. Learned semantic localization~\cite{shorinwa2024splat} and LLM priors~\cite{wang2023robogen} can inform object and object configuration randomization in an environment. Certain object classes can be generated heuristically at scale~\cite{yang2025deep}. In addition, for a known scenario, an oracle planner in a simulated environment can generate new trajectories within demonstrator-annotated~\cite{oh2026self} or task-specific~\cite{zhou2023nerf, ke2021grasping} precision from the original, generate new paths around detected 3D obstacles~\cite{pan2025one}, and retarget demonstrations to perturbed object configurations either in a simulator~\cite{jiang2025dexmimicgen, mandlekar2023mimicgen} or in an edited 3D point cloud~\cite{xue2025demogen}. We propose a method to expand expert demonstrations to a more diverse set of trajectories without requiring an oracle policy or additional inductive biases. Our proposed data augmentation method generates recovery trajectories targeting areas of the state space that the policy would likely visit due to errors in following the expert data. We show that this targeted augmentation generated after calibrating with the policy's error performs as well as using the result of a noise level hyperparameter sweep.

\section{Preliminaries}

\subsection{Noise Injection for Robust Imitation Learning}

We aim to minimize covariate shift by optimizing the KL-divergence between the trajectories visited by the trained robot policy $p(\xi | \pi_{\theta})$ and those visited by the noised expert $p(\xi | \pi_{\theta^*}, \psi)$.



In DART~\cite{laskey2017dart}, the authors minimize covariate shift by optimizing the KL-divergence with respect to the sufficient statistics $\psi$ of the noise distribution. They derive an iterative algorithm to compute $\Sigma^{\alpha}_{k+1}$ to estimate the noise parameter of a Gaussian with covariance $\Sigma$:

\begin{equation}
\begin{split}
        \quad \hat{\Sigma}_{k+1}=\frac{1}{T}& E_{p\left(\xi \mid \pi_{\theta^*}, \Sigma_k^\alpha\right)} \\
        &\sum_{t=0}^{T-1}\left(\pi_{\theta}\left(s_t\right)-\pi_{\theta^*}\left(s_t\right)\right)\left(\pi_{\theta}\left(s_t\right)-\pi_{\theta^*}\left(s_t\right)\right)^T
\end{split}
\label{eq:dart-noise-estimation-part1}
\end{equation}

\begin{equation}
    \Sigma_{k+1}^\alpha=\frac{\alpha}{T \operatorname{tr}\left(\hat{\Sigma}_{k+1}\right)} \hat{\Sigma}_{k+1}
    \label{eq:dart-noise-estimation}
\end{equation}

This requires querying the expert at noised states $\xi \sim p\left(\xi \mid \pi_{\theta^*}, \Sigma_k^\alpha\right)$ in closed loop, which is labor-intensive. In addition, the estimate of the robot's final error $\alpha$ requires hyperparameter tuning, so the process must be repeated. 


\section{Policy-Calibrated DAgger}

We introduce Policy-Calibrated DAgger, a method that predicts data augmentation noise parameters that make the expert demonstrations cover states closer to the final robot’s distribution. A key difference between our method and DART~\cite{laskey2017dart} is that we do not inject noise into the expert’s control loop—instead, we compute the sufficient statistics offline and in a simulator. This places the burden of corrections on the policy instead of the expert and allows the expert to provide clean demonstrations, uninterrupted.

\subsection{Method Design}

We present a method that uses policy-calibrated data augmentations in the DAgger training loop. We generate targeted recovery trajectories that match policy failures during evaluation using a noise metric computed entirely offline by exploiting known noise schedules in generative models. Our system is designed as follows, as shown in~\autoref{fig:method_overview}: 


\begin{enumerate}
    \item Train a policy with imitation learning using some base dataset of expert demonstrations $\mathcal{D}_0$.
    \item Aggregate expert intervention data from the $k$th round of DAgger $\mathcal{D}_k$ into the training dataset to create the aggregate dataset $\mathcal{D}_{agg,k} = \mathcal{D}_0 \cup \mathcal{D}_1 \cup \ldots \cup \mathcal{D}_k$.
    \item Generate augmented intervention data $\hat{\mathcal{D}}_k$ from the intervention data $\mathcal{D}_k$ by noising states from the recorded expert intervention data $\mathcal{D}_k$ and generating a trajectory that returns to the expert trajectory. See~\autoref{ssec:estimating-policy-calibration-parameters} for details. This creates the aggregate augmented dataset $\hat{\mathcal{D}}_{agg,k} = \hat{\mathcal{D}}_1 \cup \hat{\mathcal{D}}_2 \cup \ldots \cup \hat{\mathcal{D}}_k$ (created on the fly in the training loop in practice).
    \item Co-train the policy on the aggregate dataset $\mathcal{D}_{agg,k}$ and the augmented intervention data $\hat{\mathcal{D}}_{agg,k}$.
    \item Repeat steps 2-4 for the number of dataset aggregation rounds.
    \item For final evaluation statistics, we fine-tune Step 1’s policy $\pi_0$ on all rounds' aggregate and noised datasets.
\end{enumerate}

To collect intervention data, we follow HG-DAgger~\cite{kelly2019hg} and collect expert-gated interventions. This is in contrast to DAgger, which switches between the policy and the expert at each timestep and may make it hard for the expert to provide demonstrations. For reproducibility in our experiments, we do not use human expert data but instead use a deterministic RRT-based oracle. The oracle uses privileged information in a simulator to decide when to return control back to the policy based on heuristics like whether the policy's predicted action chunk will drive the robot into collision or if the policy has not yet reached the goal after a given number of timesteps.

\begin{figure}[t]
    \centering
    \includegraphics[width=0.9\linewidth]{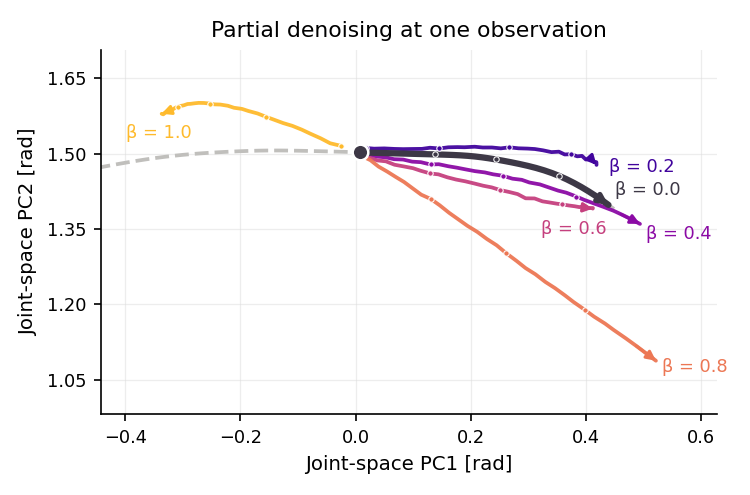}
    \caption{Empirically, performing partial denoising at lower blend ratios $\beta$ results in action chunks closer to the sampled expert chunk. The unguided policy $\beta=1.0$ may sometimes travel in a different direction than the sampled expert chunk.}
    \label{fig:partial-denoise-at-one-spot}
\vspace{-0.3cm}
    
\end{figure}

\subsection{Partial Denoising as a Learned Action Interpolator}
\label{ssec:partial-diffusion}

We leverage a generative policy's properties to model its deviation from the expert data. Because the policy may solve the task in a different way from the expert, making common error metrics ill-posed, we interpolate between the expert and policy actions with partial denoising. Then, we use the policy's noise schedule to recover an estimate of the policy's true error with respect to the expert data.


At state $s_t$, the robot policy $\pi_\theta$ sees an observation $o_t \in~\mathbb{R}^n$ and predicts action $a_{\theta,t} \in \mathbb{R}^d$ at timestep $t$. The expert prefers the action $a_{\theta^*,t} \in \mathbb{R}^d$ at state $s_t$. A common method for blending actions between an expert and a policy is linear interpolation~\cite{jonnavittula2024sari, zhou2026saps}. With linear interpolation, the blend ratio $\beta \in [0, 1]$ controls the extent to which the expert's action is preserved relative to the robot action in the blended action $\hat{a}_{t,\text{lerp}}$, where $\beta=0$ is fully the expert action and $\beta=1$ is fully the policy action:

\vspace{-0.3cm}

\begin{equation}
    \hat{a}_{t,\text{lerp}} = \beta a_{\theta,t} + (1 - \beta) a_{\theta^*,t}
\end{equation}

Linear interpolation  can fail to produce a valid action even if both $a_{\theta, t}$ and $a_{\theta^*, t}$ are valid. For example, imagine a scenario where the robot must avoid an obstacle ahead of it. Also, assume that $a_{\theta, t}$ represents avoiding the obstacle by going to the left, and $a_{\theta^*, t}$ represents avoiding the obstacle by going to the right. The blended action $\hat{a}_t$ at $\beta = 0.5$ would crash the robot into the obstacle by going forwards. A learned interpolator may project the action space in a way that the middle of the two actions is semantically meaningful.

We can leverage learned priors in the policy $\pi_{\theta}$ to blend $a_{\theta^*,t}$ and $a_{\theta,t}$ in a semantically meaningful way:

\begin{equation}
    \hat{a}_{t,\theta} = f_{\theta}(a_{\theta^*,t}, \beta | o_t) 
\end{equation}

Here, the policy parameters $\theta$ implicitly encode the robot action $a_{\theta,t}$.

One method of blending with learned priors is partial denoising~\cite{yoneda2023noise, meng2021sdedit}. Here, we will describe partial denoising for a diffusion model. A diffusion model is a generative model with a forward diffusion process and a reverse diffusion process~\cite{ho2020denoising}. The forward diffusion process adds noise to a sample from a data distribution and the reverse diffusion process denoises a noisy input. One step in the forward diffusion process can be written as:

\vspace{-0.3cm}

\begin{equation}
    a_{\theta^*, t, k} = \sqrt{\alpha_k} a_{\theta^*, t, k-1}  + \sqrt{1-\alpha_k} \epsilon, \epsilon \sim \mathcal{N}(0, I)
\end{equation}

\noindent for a noise schedule of $K$ steps $\beta_1, \ldots, \beta_K$ and $\alpha_k \coloneqq 1-\beta_k$.

The reverse diffusion process is derived as:

\vspace{-0.3cm}

\begin{equation}
    a_{\theta^*, t, k-1} = \frac{1}{\sqrt{\alpha_{k}}} (a_{\theta^*, t, k} - \frac{\beta_k}{\sqrt{1-\bar{\alpha}_k}} \epsilon_{\theta}(a_{\theta^*,t,k}, k|o_t)) + \sigma_k z
\end{equation}

\noindent where $\bar{\alpha}_k \coloneqq \prod^{k}_{s=1} \alpha_s$, $z \sim \mathcal{N}(0, I)$, and $\beta_k$ are the diffusion noise schedule coefficients.

Partial diffusion is performed as follows for a forward diffusion ratio $\beta \in [0, 1]$: 1) The switching timestep is $k_{sw} = \text{ToInteger}(\beta K)$, 2) We run the forward diffusion process for $k_{sw} - 1$ timesteps with the expert action $a_{\theta^*,t}$ as input, 3) We run the reverse diffusion process for the same $k_{sw} - 1$ timesteps on the result of the second step. More details can be found in~\cite{yoneda2023noise}.

The extent to which the original action is preserved is bounded by the number of forward and reverse diffusion process steps performed. More formally, as derived in~\cite{meng2021sdedit}, assuming for the predicted noise $\epsilon_{\theta}(a, k | o)$ that $||\epsilon_{\theta}(a, k | o)|| \leq \mathcal{K} \text{ } \forall a \in \mathcal{A}$ for $k \in [0, K]$ and actions $a \in [-1, 1]$ are normalized, then for all $\delta \in (0, 1)$ with probability at least $(1-\delta)$: 

\vspace{-0.3cm}

\begin{equation}
    || a_{\theta^*,t} - \hat{a}_{t, \theta} ||^2_2 \leq \sigma^2_{k_{sw}}(\mathcal{K}\sigma^2_{k_{sw}} + d + 2\sqrt{-d \cdot \log\delta} - 2\log\delta)
    \label{eq:partial-diffusion-bound}
\end{equation}

\noindent where $d$ is the dimensionality of $a$. Empirically, it has been shown that a larger number of diffusion steps $k_{sw}$ results in larger differences between the original action $a_{\theta^*, t}$ and the action resulting from partial diffusion $\hat{a}_{t, \theta}$~\cite{yoneda2023noise}. See \autoref{fig:partial-denoise-at-one-spot}.

\subsection{Extending Partial Denoising to Action Chunks}

We extend partial denoising to action chunks $\overrightarrow{\boldsymbol{a}}_{\theta^*,t:t+H} \in \mathbb{R}^{d \times H}$ of time horizon $H$. Theoretically, the method remains the same because we can flatten the $\overrightarrow{\boldsymbol{a}}_{\theta^*,t:t+H}$ matrix into a vector. In practice, this means we must request actions from the expert that span some time into the future. We circumvent this by processing expert data offline and leave the problem of expert intent inference to future work. The use of action chunks means that an expert must intervene for at least $H$ time steps for the policy to be able to learn from the data. We discard interventions shorter than $H$ timesteps. 

\subsection{Estimating Policy Calibration Parameters}
\label{ssec:estimating-policy-calibration-parameters}

We propose methods to empirically measure the unknown values in~\autoref{eq:dart-noise-estimation} in simulation, without requiring an expert in the loop.

First, we estimate the unknown $\hat{\Sigma}_{k+1}$ in~\autoref{eq:dart-noise-estimation-part1} by sampling a segment of a recorded expert trajectory. We approximate $\hat{\Sigma}_{k+1}$ by sampling from the expert trajectories $\xi \sim p(\xi~\mid~\pi_{\theta^*})$ instead of from the noised expert trajectories $\xi \sim p\left(\xi \mid \pi_{\theta^*}, \Sigma_k^\alpha\right)$ because we find that it results in a $\hat{\Sigma}_{k+1}$ close to the true value in practice. This approximation lets us sample open loop, without a simulator. We set the policy's observations to the beginning of a trajectory segment then compare the policy's predicted action chunk with the expert's future trajectory. From~\autoref{eq:dart-noise-estimation-part1}, $\pi_{\theta}\left(s_t\right)$ is the policy's predicted action chunk and $\pi_{\theta^*}\left(s_t\right)$ is the expert trajectory.

We also estimate the robot's final closed-loop error $\alpha$ from~\autoref{eq:dart-noise-estimation}. One method to measure $\alpha$ in simulation is to roll out the policy initialized from the expert trajectory then measure the accumulated error over time. However, a suboptimal policy tends to quickly diverge from the expert without supervision. This may be from the multimodality of the task. Large deviations make measurements like the L2 norm ill-posed. We instead introduce a method to retain some of the policy while making it more likely for sampled trajectories to stay close to the expert intervention. To do this, we simulate the equilibrium of two competing forces: keeping the trajectory close to the expert trajectory and pulling the trajectory towards the policy's true error. 

Let $x \in \mathbb{R}^d$ represent the state error relative to the expert trajectory. From~\autoref{eq:partial-diffusion-bound}, the measured error between a sample of partial diffusion $\hat{a}_{t,\theta}$ and the expert data $a_{\theta^*,t}$ is bounded by the maximum policy predicted noise $\mathcal{K}$ scaled by $\sigma^2_{k_{sw}}$, where the switching step in partial diffusion $k_{sw} = \text{ToInteger}(\beta K)$. The noise schedule at the switching timestep of partial denoising for a diffusion model is $\sigma_{k_{sw}}^2 = 1 - \bar{\alpha}_{k_{sw}} \in [0, 1]$. We use $\sigma_{k_{sw}}^2$ to model the interaction between the expert and the policy:

\vspace{-0.4cm}
\begin{equation}
    u_{\text{blend}} = \sigma_{k_{sw}}^2 \cdot u_{\theta}  + (1-\sigma_{k_{sw}}^2) \cdot u_{\theta^*}
\end{equation}

For an expert $u_{\theta^*}$ and policy $u_{\theta}$, we assume the expert settles at $0$ error and has an isotropic gain $ K_{\theta^*}>0$ for all dimensions $d$:

\vspace{-0.4cm}

\begin{equation}
    u_{\theta^*} = K_{\theta^*} \cdot (-x)
\end{equation}

\noindent And we assume the policy settles towards an alternate goal $W$ with isotropic gain $ K_{\theta}>0$:

\vspace{-0.3cm}

\begin{equation}
    u_{\theta} = K_{\theta} \cdot (W - x)
\end{equation}

At the equilibrium, $u_{\text{blend}} = 0$ so 
\begin{equation}
     0 = \sigma_{k_{sw}}^2 \cdot u_{\theta}  + (1-\sigma_{k_{sw}}^2) \cdot u_{\theta^*}
\end{equation}

\begin{equation}
     x = \frac{W \cdot \sigma_{k_{sw}}^2}{\sigma_{k_{sw}}^2 + \frac{ K_{\theta^*}}{ K_{\theta}}(1-\sigma_{k_{sw}}^2)}
     \label{eq:blend-equilibrium}
\end{equation}

\begin{figure}[!t]
    \centering
    \includegraphics[width=0.95\linewidth]{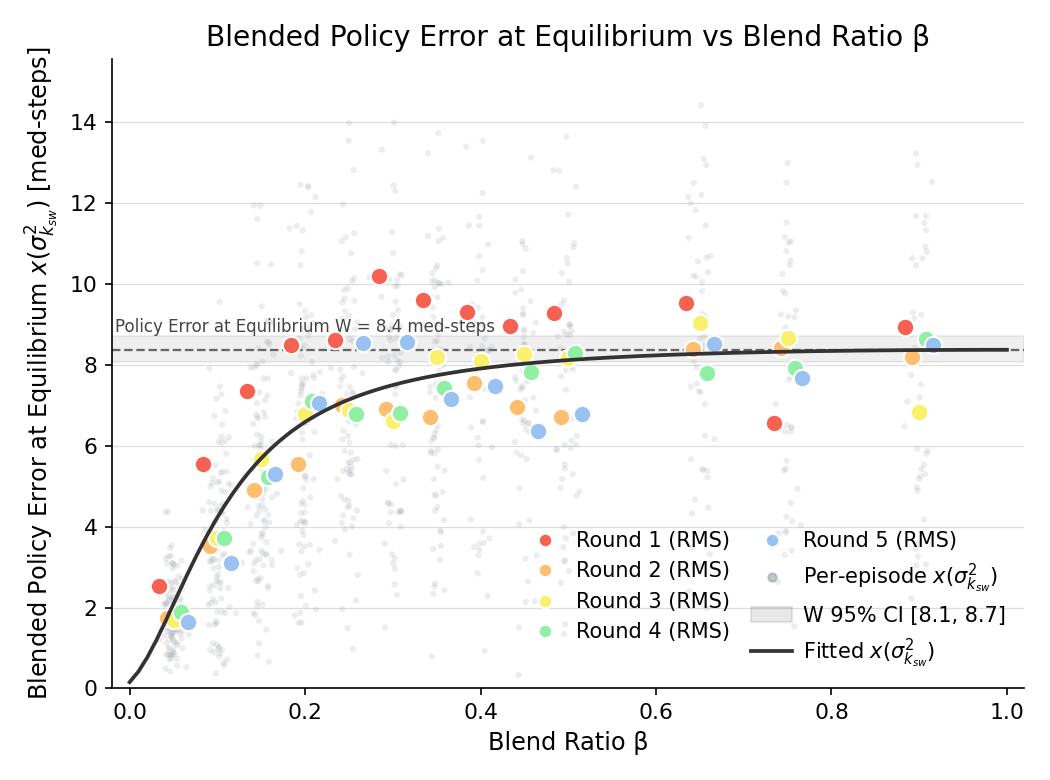}
    \caption{Our fitted curve for estimating $W$ on the planar reacher environment. As blend ratio $\beta \rightarrow 1$, noise variance fraction $\sigma_{k_{sw}}^2 \rightarrow 1$ and thus our estimate $x(\sigma_{k_{sw}}^2) \rightarrow W$. We measure error in terms of median step size measured from the demonstration data.}
    \label{fig:blend-fit-for-w}
    \vspace{-0.5cm}
\end{figure}

To recover the unknowns $W$ and $\frac{ K_{\theta^*}}{ K_{\theta}}$, we sample the policy deviations after rolling out a ``blended" trajectory for multiple blend ratios, where we compute partial denoising with respect to the time-matched future expert trajectory at each timestep, then execute the resulting blend action. We are interested in the settling state at equilibrium, so we discard the blended trajectories that diverge from the expert trajectory. Divergence is more likely at high blend ratios $\beta$. Between training iterations, we sample 13 blend ratios in [0.05, 0.90] once each and fit~\autoref{eq:blend-equilibrium}. Although $W$ is the settling error of the current policy, not of the final policy, we find that $W$ tends to be stable across rounds $k$. See~\autoref{fig:blend-fit-for-w}.

An approximation of DART's $\alpha$ with the measured settling error $W$ is:
\begin{equation}
    \alpha = T ||W||_2^2
\end{equation}

And thus:
\begin{equation}
    \hat{\Sigma}_{k+1}^\alpha = \frac{||W||_2^2}{\operatorname{tr}\left(\hat{\Sigma}_{k+1}\right)} \hat{\Sigma}_{k+1}
\end{equation}

Since we computed $\hat{\Sigma}_{k+1}$ at every timestep in the expert data, $\Sigma^{\alpha}_{k+1}$ is a per-step measurement of error. We compute a pooled noise measurement to reduce sample noise and also better match the per-round constant in DART~\cite{laskey2017dart}. For each of $m \in \mathcal{M}_{k+1}$ measurements of $\hat{\Sigma}_{k+1}$:

\vspace{-0.3cm}
\begin{equation}
    \bar{\Sigma}_{k+1} = \frac{1}{|\mathcal{M}_{k+1}|} \sum_{m \in \mathcal{M}_{k+1}} \hat{\Sigma}_{m}
\end{equation}

\begin{equation}
    \bar{\Sigma}^{\alpha}_{k+1} = \frac{||W||_2^2}{\text{tr}(\bar{\Sigma}_{k+1})}\bar{\Sigma}_{k+1}
\end{equation}

Our policy calibration method is in~\autoref{fig:method_overview}.

\subsection{Generating Corrective Labels}

We generate state-action pairs within the noise distribution that encourage the policy to return to the known data distribution. We assume access to a heuristic algorithm that connects a pair of nearby states. Action generation is done with respect to the demonstration dataset's velocity and acceleration statistics. For a perturbed starting state from our noise distribution, we label its action as an action chunk that starts from the perturbed state and returns to the original expert intervention trajectory. These action chunks can be generated cheaply because our policy's actions are joint positions. We ensure that these generated corrective action chunks are smooth and in distribution with the original dataset. We first associate the augmented state to a timestep of the associated expert intervention trajectory with a monotone closest-point projection. Then, we create a smooth trajectory between the augmented state and a state a few timesteps further in the associated expert trajectory state (computed with the median velocity in the dataset) by rolling out a critically-damped PD controller, with accelerations clamped to those in the original dataset's demonstrations.

\begin{table*}[ht]
\vspace{0.2cm}
\centering
\setlength{\extrarowheight}{2pt} 
\begin{tabular}{l|c|c|c|c|c}
\hline
Method & $K{=}1$ & $K{=}2$ & $K{=}3$ & $K{=}4$ & $K{=}5$ \\
\hline
BC (no aggregation) & \multicolumn{5}{c}{68.0 $\pm$ 0.2\textsuperscript{(1)}} \\
\hline
HG-DAgger~\cite{kelly2019hg} & \cellcolor[HTML]{E2EDF4}79.9 $\pm$ 0.5\textsuperscript{(5)} & \cellcolor[HTML]{DEEBF3}81.9 $\pm$ 0.7\textsuperscript{(5)} & \cellcolor[HTML]{C2E0F1}83.3 $\pm$ 1.5\textsuperscript{(5)} & \cellcolor[HTML]{AED7EF}85.3 $\pm$ 0.7\textsuperscript{(5)} & \cellcolor[HTML]{99CFEE}87.3 $\pm$ 1.0\textsuperscript{(5)} \\
\noalign{\hrule height 0.9pt}
Fixed iso.\ noise ($\sigma{=}2$)~\cite{ke2021grasping} & \cellcolor[HTML]{A8D5EF}81.9 $\pm$ 0.8\textsuperscript{(5)} & \cellcolor[HTML]{8FCBED}84.8 $\pm$ 0.7\textsuperscript{(5)} & \cellcolor[HTML]{8AC9ED}85.1 $\pm$ 0.3\textsuperscript{(5)} & \cellcolor[HTML]{5DB6E9}88.6 $\pm$ 0.4\textsuperscript{(5)} & \cellcolor[HTML]{6BBCEA}89.1 $\pm$ 0.4\textsuperscript{(5)} \\
Fixed iso.\ noise ($\sigma{=}4$) & \cellcolor[HTML]{8FCAED}82.7 $\pm$ 0.9\textsuperscript{(5)} & \cellcolor[HTML]{6DBCEA}86.1 $\pm$ 0.7\textsuperscript{(5)} & \cellcolor[HTML]{66BAEA}86.3 $\pm$ 0.5\textsuperscript{(5)} & \cellcolor[HTML]{73BFEB}87.7 $\pm$ 0.6\textsuperscript{(5)} & \cellcolor[HTML]{68BAEA}89.2 $\pm$ 0.8\textsuperscript{(5)} \\
Fixed iso.\ noise ($\sigma{=}8$) & \cellcolor[HTML]{6DBCEA}83.9 $\pm$ 0.8\textsuperscript{(5)} & \cellcolor[HTML]{88C8EC}85.1 $\pm$ 0.8\textsuperscript{(5)} & \cellcolor[HTML]{8CC9ED}85.1 $\pm$ 0.8\textsuperscript{(5)} & \cellcolor[HTML]{9ACFEE}86.1 $\pm$ 0.8\textsuperscript{(5)} & \cellcolor[HTML]{A0D1EE}87.1 $\pm$ 0.8\textsuperscript{(5)} \\
Fixed iso.\ noise ($\sigma{=}12$) & \cellcolor[HTML]{7DC3EB}83.3 $\pm$ 0.7\textsuperscript{(5)} & \cellcolor[HTML]{8FCBED}84.8 $\pm$ 0.6\textsuperscript{(5)} & \cellcolor[HTML]{8AC9ED}85.1 $\pm$ 0.8\textsuperscript{(5)} & \cellcolor[HTML]{C0DFF1}84.5 $\pm$ 1.0\textsuperscript{(5)} & \cellcolor[HTML]{D6E8F3}85.0 $\pm$ 0.8\textsuperscript{(5)} \\
Fixed iso.\ noise ($\sigma{=}16$) & \cellcolor[HTML]{81C5EC}83.2 $\pm$ 1.3\textsuperscript{(5)} & \cellcolor[HTML]{86C7EC}85.1 $\pm$ 0.6\textsuperscript{(5)} & \cellcolor[HTML]{B2D9F0}83.8 $\pm$ 0.9\textsuperscript{(5)} & \cellcolor[HTML]{DBEAF3}83.4 $\pm$ 0.9\textsuperscript{(5)} & \cellcolor[HTML]{DBEAF3}84.8 $\pm$ 0.8\textsuperscript{(5)} \\
\noalign{\hrule height 0.9pt}
Pooled $\bar{\Sigma}^{\alpha}$ (Ours) & \cellcolor[HTML]{A5D4EF}82.0 $\pm$ 0.6\textsuperscript{(5)} & \cellcolor[HTML]{69BBEA}86.2 $\pm$ 0.8\textsuperscript{(5)} & \cellcolor[HTML]{60B7E9}86.5 $\pm$ 0.4\textsuperscript{(5)} & \cellcolor[HTML]{78C1EB}87.5 $\pm$ 0.9\textsuperscript{(5)} & \cellcolor[HTML]{64B9E9}89.3 $\pm$ 0.5\textsuperscript{(5)} \\
Per-step $\hat\Sigma^{\alpha}$ (Ours) & \cellcolor[HTML]{B0D9F0}81.6 $\pm$ 0.9\textsuperscript{(5)} & \cellcolor[HTML]{82C5EC}85.3 $\pm$ 0.9\textsuperscript{(5)} & \cellcolor[HTML]{80C4EC}85.5 $\pm$ 0.3\textsuperscript{(5)} & \cellcolor[HTML]{8FCBED}86.5 $\pm$ 0.5\textsuperscript{(5)} & \cellcolor[HTML]{79C2EB}88.5 $\pm$ 0.8\textsuperscript{(5)} \\
\noalign{\hrule height 0.9pt}
Closed-loop pooled $\bar{\Sigma}^{\alpha}$ (Ours) & \cellcolor[HTML]{B2D9F0}81.6 $\pm$ 1.0\textsuperscript{(1)} & \cellcolor[HTML]{B6DBF0}83.3 $\pm$ 0.2\textsuperscript{(1)} & \cellcolor[HTML]{74BFEB}85.9 $\pm$ 0.5\textsuperscript{(1)} & \cellcolor[HTML]{58B4E8}88.8 $\pm$ 0.2\textsuperscript{(1)} & \cellcolor[HTML]{64B9E9}89.3 $\pm$ 0.7\textsuperscript{(1)} \\
\hline
\end{tabular}
\caption{Mean$\pm$SE success rates on a 100-scenario planar reacher environment (3 evaluations per scenario). Most cells are averaged across 5 DAgger lineages$^{(5)}$. BC is trained on $\mathcal{D}_0$ (500 demonstrations). To prevent sequential optimization errors, all policy variants are fine-tuned from BC's 75k checkpoint for each round $K$ (175k total steps). We sweep over isotropic noise levels and calibrated noise estimated before training for the round. Darker colors denote higher success. Our method matches the performance of the best noise level in the hyperparameter sweep without requiring extra expert data.}
\label{tab:overall_success}

\vspace{-0.5cm}

\end{table*}

\begin{table}[t]
\centering
\setlength{\extrarowheight}{2pt} 
\begin{tabular}{l|c|c}
\hline
Method & $K{=}1$ & $K{=}2$\\
\hline
BC (no aggregation) & \multicolumn{2}{c}{58.0 $\pm$ 1.2\textsuperscript{(1)}} \\
\hline
HG-DAgger~\cite{kelly2019hg} & 57.4 $\pm$ 5.6\textsuperscript{(3)} & 58.1 $\pm$ 4.0\textsuperscript{(3)} \\
\noalign{\hrule height 0.9pt}
Pooled $\bar{\Sigma}^{\alpha}$ (Ours) & 68.8 $\pm$ 4.2\textsuperscript{(3)} & 63.4 $\pm$ 4.8\textsuperscript{(3)}\\
\hline
\end{tabular}
\caption{Mean$\pm$SE success rates on a 100-scenario photorealistic engine lever-reaching environment (3 evaluations per scenario). Following~\autoref{tab:overall_success}'s protocol, policy variants are fine-tuned from BC's 75k training step checkpoint to a total of 95k training steps. BC is trained on 500 demonstrations.}
\label{tab:overall_success_splatsim}
\vspace{-0.4cm}
\end{table}

\begin{figure}[t]
    \centering
    \includegraphics[width=0.97\linewidth]{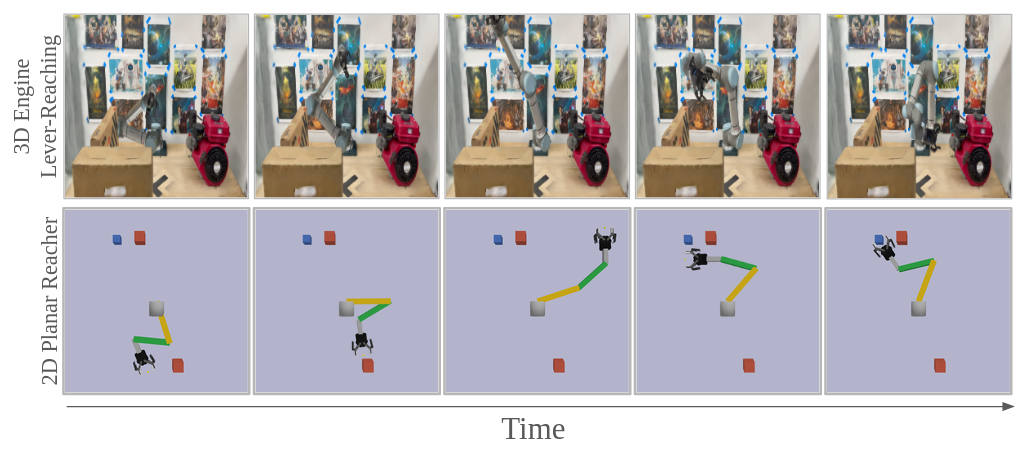}
    \caption{We evaluate our method in two cluttered goal-reaching tasks with narrow passageways, where the robot often must untangle its links to make progress towards the goal. The 3D photorealistic physics simulation recreates our real mockup. We run ablations on the 3-link planar robot.}
    \label{fig:env_example}
    \vspace{-0.6cm}
    
\end{figure}

\section{Experiments}
\label{sec:experiments}

    \vspace{-0.2cm}

We conduct a series of experiments to validate our method for data-efficient interactive imitation learning. Our primary research question is how our proposed augmentations to intervention data affect policy optimization given the same number of expert interventions. Our augmentations are based on a diffusion model's~\cite{chi2025diffusion} denoising process. 
Specifically, we compare policy performance among a few data collection and learning methods: \textbf{(1)} Offline Behavior Cloning (BC), \textbf{(2)} Expert-gated data collection where the expert determines when to intervene and release control back to the policy (HG-DAgger~\cite{kelly2019hg}), \textbf{(3)} (2) but with a few fixed noise levels~\cite{ke2021grasping}, and \textbf{(4)} (2) but noised with our predicted noise level at each round (Ours).
Through these experiments, our aim is to demonstrate that data augmentations that match the distribution of errors that the policy will induce during evaluation time can improve interactive imitation learning without additional data collection.



We build our own simulator because existing benchmarks either lack an expert that can correct a policy from arbitrary states or lack the clutter needed to expose compounding error. We experiment on an engine lever-reaching task, where a UR5 manipulator arm navigates a cluttered, narrow passageway with randomized obstacles to reach an engine lever. All robot joints must move precisely to avoid collision, since an end effector-based action space would be underconstrained. We recreate a real engine lever-reaching task in a photorealistic simulation by following SplatSim~\cite{qureshi2025splatsim}. We scan our physical experimental setup and objects with a smartphone to train Gaussian splats~\cite{kerbl20233d}. Then, we associate links in the robot's URDF with points in its splat. When a URDF moves in the PyBullet~\cite{coumans2016pybullet} physics simulation, we transform the associated points in its splat. In addition, we run ablations on a 2D planar reacher environment where a 3-link robot is tasked with reaching a goal end effector position while avoiding obstacles. Due to the clutter, the robot's embodiment often becomes the main collision obstacle in both tasks, and the robot must sometimes take a roundabout route to untangle its joints and links.

To improve reproducibility, we use a RRT-based oracle expert to generate intervention data in the loop when the evaluated policy is close to collision or stalled. We use this expert for both environments. The oracle plans a path from the current state to the goal, but we execute only a fraction of the plan before passing control back to the policy. This emulates how a human expert drives the robot just enough to return it to a safe state in human-gated DAgger. The oracle expert takes the optimal path under a deterministic criterion (for example, minimum end effector arc length). In addition, we smooth the RRT-generated trajectory with random shortcutting, trajectory optimization, corner smoothing,
and speed, acceleration, and jerk limits enforced by TOPP-RA~\cite{pham2018new}. The path planner constrains the generated path to match the policy's induced pose and velocity at the handoff. 

We conduct paired experiments where we first train a diffusion policy on the base dataset of 500 demonstrations on randomized scenes, then iteratively collect intervention data and fine-tune the policy without using any noised data. Our results for the planar reacher task success rates across method configurations are shown in~\autoref{tab:overall_success}. Our noised policies are fine-tuned from a checkpoint of the BC policy on the aggregate of: the base dataset $\mathcal{D}_0$, intervention data up to the given round of DAgger $\mathcal{D}_1\ldots\mathcal{D}_K$, and noise augmentations applied to the intervention data $\hat{\mathcal{D}}_1\ldots\hat{\mathcal{D}}_K$. This means that our noised policies are paired with the HG-DAgger results that do not use noise during training or data collection. In addition, the noised policies show performance gains over HG-DAgger and BC without having access to data from a noised policy. We present results where each policy is fine-tuned for the same number of steps within each environment, starting from the same checkpoint trained on the base dataset.

For the planar reacher task, BC without any dataset aggregation performs the worst at 68.0\% success. The baseline data aggregation method HG-DAgger~\cite{kelly2019hg} increases the policy success rate from 79.9\% to 87.3\% success over five data collection iterations. In addition, most noised variants further increase the success rate over using $\mathcal{D}_k$ alone. Averaged over rounds and paired across lineages, pooled calibrated noise improves on HG-DAgger by 2.8 points ($t(4) = 5.75$, $p = 0.005$) and is equivalent within a $\pm 2$ point margin to the best fixed noise level chosen in hindsight (TOST, $p \leq 0.01$).

In the 3D photorealistic engine lever-reaching environment, we evaluate our method on a policy with image observations. As an approximation, noised observations reuse the original image observation and only the joint position labels are noised. In~\autoref{tab:overall_success_splatsim}, BC obtains a 58.0\% success rate. HG-DAgger performs similarly to BC despite additional intervention data: most failures are collisions happening within the first second of the episode, so intervention coverage is limited. In contrast, our proposed Pooled $\bar\Sigma^{\alpha}$ method scores higher than both BC and HG-DAgger when it has access to the same intervention data.

The optimal noise level tends to decrease as the policy gains competence through HG-DAgger rounds. In all five lineage reruns (collecting intervention data from scratch and seeding policy training), the best $\sigma$ is 8--16 in $K=1$ and is 2--4 by $K=4$. Large noise levels like $\sigma = 12 \text{ and } 16$ hurt the success rate in late rounds $K=4\text{ and }5$, while small noise levels like $\sigma=2$ show an average 2 point increase in success in all DAgger rounds. Our method performs similarly to the best fixed isotropic noise ratio, but without sweeping over noise magnitudes. Our measured $\bar\Sigma^{\alpha}$ has a per-joint $\sigma$ of 3.9--5.1 median per-step joint displacement in 2D planar reacher, close to the best-performing isotropic noise level $\sigma=4$, and has a per-joint $\sigma$ of 1.3--4.2 median steps in 3D engine lever-reaching in which the robot moves at similar velocities as the 2D planar reacher environment.

Another insight is that augmenting action labels may provide benefit in addition to only observation augmentations. Across all rounds $k$ in the planar environment, even the smallest noise level in the sweep ($\sigma=2$) increases success rate compared to HG-DAgger. This action relabeling was done in addition to the common practice of applying small Gaussian noise to the observations without changing action labels to simulate inaccurate signal readings.


\vspace{-0.1cm}

\section{Conclusion}

\vspace{-0.1cm}

We present Policy-Calibrated DAgger, a method that applies policy-calibrated noise to data collected during iterative imitation learning, producing more robust policies without requiring additional expert supervision. Our method achieves higher success rates than behavior cloning and HG-DAgger, and matches performance with the best noise level in hindsight, without an expensive hyperparameter sweep over noise levels with an expert in the loop. Instead, we trade expert time for compute. We show that these trajectory data augmentations target policy errors during evaluation time and the increased success rates are consistent with reduced covariate shift. We use the policy as its own measure of noise, making our method agnostic to the semantics of individual action dimensions. Policy error is ill-defined when the trajectory to measure error with respect to does not exist in the policy's representation; however, we find that this gap is small in practice. Future work could derive partial denoising for other generative models such as flow matching models, apply our method to human-in-the-loop data collection, and extend our method to online noise estimation.

\vspace{-0.2cm}

\bibliographystyle{IEEEtran}
\bibliography{main_bib}

@article{liu2022revolver,
  title={Revolver: Continuous evolutionary models for robot-to-robot policy transfer},
  author={Liu, Xingyu and Pathak, Deepak and Kitani, Kris M},
  journal={arXiv preprint arXiv:2202.05244},
  year={2022}
}

@inproceedings{tao2025incremental,
  title={Incremental learning for robot shared autonomy},
  author={Tao, Yiran and Qiao, Guixiu and Ding, Dan and Erickson, Zackory},
  booktitle={2025 IEEE/RSJ International Conference on Intelligent Robots and Systems (IROS)},
  pages={20797--20804},
  year={2025},
  organization={IEEE}
}

@article{yoneda2023noise,
  title={To the noise and back: Diffusion for shared autonomy},
  author={Yoneda, Takuma and Sun, Luzhe and Yang, Ge and Stadie, Bradly and Walter, Matthew},
  journal={arXiv preprint arXiv:2302.12244},
  year={2023}
}

@article{xu2026compliant,
  title={Compliant residual dagger: Improving real-world contact-rich manipulation with human corrections},
  author={Xu, Xiaomeng and Hou, Yifan and Liu, Zeyi and Song, Shuran},
  journal={Advances in Neural Information Processing Systems},
  volume={38},
  pages={139559--139581},
  year={2026}
}

@article{wang2023robogen,
  title={Robogen: Towards unleashing infinite data for automated robot learning via generative simulation},
  author={Wang, Yufei and Xian, Zhou and Chen, Feng and Wang, Tsun-Hsuan and Wang, Yian and Fragkiadaki, Katerina and Erickson, Zackory and Held, David and Gan, Chuang},
  journal={arXiv preprint arXiv:2311.01455},
  year={2023}
}

@article{liu2025factr,
  title={Factr: Force-attending curriculum training for contact-rich policy learning},
  author={Liu, Jason Jingzhou and Li, Yulong and Shaw, Kenneth and Tao, Tony and Salakhutdinov, Ruslan and Pathak, Deepak},
  journal={arXiv preprint arXiv:2502.17432},
  year={2025}
}

@article{wu2025robocopilot,
  title={Robocopilot: Human-in-the-loop interactive imitation learning for robot manipulation},
  author={Wu, Philipp and Shentu, Yide and Liao, Qiayuan and Jin, Ding and Guo, Menglong and Sreenath, Koushil and Lin, Xingyu and Abbeel, Pieter},
  journal={arXiv preprint arXiv:2503.07771},
  year={2025}
}

@article{bazhenov2025echo,
  title={Echo: An open-source, low-cost teleoperation system with force feedback for dataset collection in robot learning},
  author={Bazhenov, Artem and Satsevich, Sergei and Egorov, Sergei and Khabibullin, Farit and Tsetserukou, Dzmitry},
  journal={arXiv preprint arXiv:2504.07939},
  year={2025}
}

@article{chi2024universal,
  title={Universal manipulation interface: In-the-wild robot teaching without in-the-wild robots},
  author={Chi, Cheng and Xu, Zhenjia and Pan, Chuer and Cousineau, Eric and Burchfiel, Benjamin and Feng, Siyuan and Tedrake, Russ and Song, Shuran},
  journal={arXiv preprint arXiv:2402.10329},
  year={2024}
}

@inproceedings{pan2025one,
  title={One Demo is Worth a Thousand Trajectories: Action-View Augmentation for Visuomotor Policies},
  author={Pan, Chuer and Liang, Litian and Bauer, Dominik and Cousineau, Eric and Burchfiel, Benjamin and Feng, Siyuan and Song, Shuran},
  booktitle={9th Annual Conference on Robot Learning},
  year={2025}
}

@inproceedings{zhou2023nerf,
  title={Nerf in the palm of your hand: Corrective augmentation for robotics via novel-view synthesis},
  author={Zhou, Allan and Kim, Moo Jin and Wang, Lirui and Florence, Pete and Finn, Chelsea},
  booktitle={Proceedings of the IEEE/CVF Conference on Computer Vision and Pattern Recognition},
  pages={17907--17917},
  year={2023}
}

@inproceedings{jiang2025dexmimicgen,
  title={Dexmimicgen: Automated data generation for bimanual dexterous manipulation via imitation learning},
  author={Jiang, Zhenyu and Xie, Yuqi and Lin, Kevin and Xu, Zhenjia and Wan, Weikang and Mandlekar, Ajay and Fan, Linxi Jim and Zhu, Yuke},
  booktitle={2025 IEEE International Conference on Robotics and Automation (ICRA)},
  pages={16923--16930},
  year={2025},
  organization={IEEE}
}

@article{oh2026self,
  title={Self-augmented robot trajectory: efficient imitation learning via safe self-augmentation with demonstrator-annotated precision},
  author={Oh, Hanbit and Murooka, Masaki and Motoda, Tomohiro and Nakajo, Ryoichi and Domae, Yukiyasu},
  journal={Advanced Robotics},
  pages={1--15},
  year={2026},
  publisher={Taylor \& Francis}
}

@inproceedings{laskey2017dart,
  title={Dart: Noise injection for robust imitation learning},
  author={Laskey, Michael and Lee, Jonathan and Fox, Roy and Dragan, Anca and Goldberg, Ken},
  booktitle={Conference on robot learning},
  pages={143--156},
  year={2017},
  organization={PMLR}
}

@inproceedings{sun2025flashback,
  title={FlashBack: Consistency Model-Accelerated Shared Autonomy},
  author={Sun, Luzhe and Ji, Jingtian and Tan, Xiangshan and Walter, Matthew},
  booktitle={Conference on Robot Learning},
  pages={924--940},
  year={2025},
  organization={PMLR}
}

@article{chen2024don,
  title={Don't start from scratch: Behavioral refinement via interpolant-based policy diffusion},
  author={Chen, Kaiqi and Lim, Eugene and Lin, Kelvin and Chen, Yiyang and Soh, Harold},
  journal={arXiv preprint arXiv:2402.16075},
  year={2024}
}

@inproceedings{kelly2019hg,
  title={Hg-dagger: Interactive imitation learning with human experts},
  author={Kelly, Michael and Sidrane, Chelsea and Driggs-Campbell, Katherine and Kochenderfer, Mykel J},
  booktitle={2019 International Conference on Robotics and Automation (ICRA)},
  pages={8077--8083},
  year={2019},
  organization={IEEE}
}

@inproceedings{ross2011reduction,
  title={A reduction of imitation learning and structured prediction to no-regret online learning},
  author={Ross, St{\'e}phane and Gordon, Geoffrey and Bagnell, Drew},
  booktitle={Proceedings of the fourteenth international conference on artificial intelligence and statistics},
  pages={627--635},
  year={2011},
  organization={JMLR Workshop and Conference Proceedings}
}

@article{liu2022herd,
  title={Herd: Continuous human-to-robot evolution for learning from human demonstration},
  author={Liu, Xingyu and Pathak, Deepak and Kitani, Kris M},
  journal={arXiv preprint arXiv:2212.04359},
  year={2022}
}

@article{maddukuri2025sim,
  title={Sim-and-real co-training: A simple recipe for vision-based robotic manipulation},
  author={Maddukuri, Abhiram and Jiang, Zhenyu and Chen, Lawrence Yunliang and Nasiriany, Soroush and Xie, Yuqi and Fang, Yu and Huang, Wenqi and Wang, Zu and Xu, Zhenjia and Chernyadev, Nikita and others},
  journal={arXiv preprint arXiv:2503.24361},
  year={2025}
}

@article{yang2025deep,
  title={Deep reactive policy: Learning reactive manipulator motion planning for dynamic environments},
  author={Yang, Jiahui and Liu, Jason Jingzhou and Li, Yulong and Khaky, Youssef and Shaw, Kenneth and Pathak, Deepak},
  journal={arXiv preprint arXiv:2509.06953},
  year={2025}
}

@article{mandlekar2023mimicgen,
  title={Mimicgen: A data generation system for scalable robot learning using human demonstrations},
  author={Mandlekar, Ajay and Nasiriany, Soroush and Wen, Bowen and Akinola, Iretiayo and Narang, Yashraj and Fan, Linxi and Zhu, Yuke and Fox, Dieter},
  journal={arXiv preprint arXiv:2310.17596},
  year={2023}
}

@article{xue2025demogen,
  title={Demogen: Synthetic demonstration generation for data-efficient visuomotor policy learning},
  author={Xue, Zhengrong and Deng, Shuying and Chen, Zhenyang and Wang, Yixuan and Yuan, Zhecheng and Xu, Huazhe},
  journal={arXiv preprint arXiv:2502.16932},
  year={2025}
}

@article{shorinwa2024splat,
  title={Splat-mover: Multi-stage, open-vocabulary robotic manipulation via editable gaussian splatting},
  author={Shorinwa, Ola and Tucker, Johnathan and Smith, Aliyah and Swann, Aiden and Chen, Timothy and Firoozi, Roya and Kennedy III, Monroe and Schwager, Mac},
  journal={arXiv preprint arXiv:2405.04378},
  year={2024}
}

@inproceedings{li2024visual,
    title={Visual robotic manipulation with depth-aware pretraining},
    author={Li, Jinming and Wang, Wanying and Peng, Yaxin and Shen, Chaomin and Zhu, Yichen and Xu, Zhiyuan},
    booktitle={2024 IEEE International Conference on Robotics and Biomimetics (ROBIO)},
    pages={843--850},
    year={2024},
    organization={IEEE}
}

@article{yu2026wmdagger,
  title={WM-DAgger: Enabling Efficient Data Aggregation for Imitation Learning with World Models},
  author={Yu, Anlan and Chen, Zaishu and Song, Peili and Hong, Zhiqing and Wang, Haotian and Zhang, Desheng and He, Tian and Ding, Yi and Zhang, Daqing},
  journal={arXiv preprint arXiv:2604.11351},
  year={2026}
}

@article{haridas2023dadagger,
  title={DADAgger: Disagreement-Augmented Dataset Aggregation},
  author={Haridas, Akash and Hamadeh, Karim and Dash, Samarendra Chandan Bindu},
  journal={arXiv preprint arXiv:2301.01348},
  year={2023}
}

@article{zhang2024diffusion,
  title={Diffusion meets dagger: Supercharging eye-in-hand imitation learning},
  author={Zhang, Xiaoyu and Chang, Matthew and Kumar, Pranav and Gupta, Saurabh},
  journal={arXiv preprint arXiv:2402.17768},
  year={2024}
}

@inproceedings{ke2021grasping,
  title={Grasping with chopsticks: Combating covariate shift in model-free imitation learning for fine manipulation},
  author={Ke, Liyiming and Wang, Jingqiang and Bhattacharjee, Tapomayukh and Boots, Byron and Srinivasa, Siddhartha},
  booktitle={2021 IEEE International Conference on Robotics and Automation (ICRA)},
  pages={6185--6191},
  year={2021},
  organization={IEEE}
}

@inproceedings{xiong2025vision,
  title={Vision in Action: Learning Active Perception from Human Demonstrations},
  author={Xiong, Haoyu and Xu, Xiaomeng and Wu, Jimmy and Hou, Yifan and Bohg, Jeannette and Song, Shuran},
  booktitle={Conference on Robot Learning},
  pages={5450--5463},
  year={2025},
  organization={PMLR}
}

@article{Khazatsky2024DROIDAL,
  title={DROID: A Large-Scale In-The-Wild Robot Manipulation Dataset},
  author={Alexander Khazatsky and Karl Pertsch and Suraj Nair and Ashwin Balakrishna and Sudeep Dasari and Siddharth Karamcheti and Soroush Nasiriany and Mohan Kumar Srirama and Lawrence Yunliang Chen and Kirsty Ellis and P Fagan and Joey Hejna and Masha Itkina and Marion Lepert and Ye Ma and Patrick Miller and Jimmy Wu and Suneel Belkhale and Shivin Dass and Huy Ha and Arhan Jain and Abraham Lee and Youngwoon Lee and Marius Memmel and Sungjae Park and Ilija Radosavovic and Kaiyuan Wang and Albert Zhan and Kevin Black and Cheng Chi and Kyle Beltran Hatch and Shan Lin and Jingpei Lu and Jean-Pierre Mercat and Abdul Rehman and Pannag R. Sanketi and Archit Sharma and C. Blake Simpson and Quang Uyển Vuong and Homer Rich Walke and Blake Wulfe and Ted Xiao and Jonathan Heewon Yang and Arefeh Yavary and Tony Zhao and Christopher Agia and Rohan Baijal and Mateo Guaman Castro and Da Ling Chen and Qiuyu Chen and Trinity Chung and Jaimyn Drake and Ethan Paul Foster and Jensen Gao and David Antonio Herrera and Minho Heo and Kyle Hsu and Jiaheng Hu and Muhammad Zubair Irshad and Donovon Jackson and Charlotte Le and Yunshuang Li and Kevin Lin and Roy Lin and Zehan Ma and Abhiram Maddukuri and Suvir Mirchandani and Daniel Morton and Tony Nguyen and Abigail O'Neill and Rosa Maria Scalise and Derick Seale and Victor Son and Stephen Tian and Emi Tran and Andrew Wang and Yilin Elaine Wu and Annie Xie and Jingyun Yang and Patrick Yin and Yunchu Zhang and Osbert Bastani and Glen Berseth and Jeannette Bohg and Ken Goldberg and Abhinav Gupta and Abhishek Gupta and Dinesh Jayaraman and Joseph J. Lim and Jitendra Malik and Roberto Mart'in-Mart'in and Subramanian Ramamoorthy and Dorsa Sadigh and Shuran Song and Jiajun Wu and Michael C. Yip and Yuke Zhu and Thomas Kollar and Sergey Levine and Chelsea Finn},
  journal={ArXiv},
  year={2024},
  volume={abs/2403.12945},
  url={https://api.semanticscholar.org/CorpusID:268531351}
}

@article{Tao2025DexWildDH,
  title={DexWild: Dexterous Human Interactions for In-the-Wild Robot Policies},
  author={Tony Tao and Mohan Kumar Srirama and Jason Jingzhou Liu and Kenneth Shaw and Deepak Pathak},
  journal={ArXiv},
  year={2025},
  volume={abs/2505.07813},
  url={https://api.semanticscholar.org/CorpusID:278534598}
}

@article{Lee2024DiffDaggerUE,
  title={Diff-Dagger: Uncertainty Estimation With Diffusion Policy for Robotic Manipulation},
  author={Sung-Wook Lee and Yen-Ling Kuo},
  journal={2025 IEEE International Conference on Robotics and Automation (ICRA)},
  year={2024},
  pages={4845-4852},
  url={https://api.semanticscholar.org/CorpusID:273502160}
}

@article{He2025UncertaintyCF,
  title={Uncertainty Comes for Free: Human-in-the-Loop Policies with Diffusion Models},
  author={Zhanpeng He and Yifeng Cao and Matei T. Ciocarlie},
  journal={ArXiv},
  year={2025},
  volume={abs/2503.01876},
  url={https://api.semanticscholar.org/CorpusID:276774935}
}

@inproceedings{Hoque2021ThriftyDAggerBN,
  title={ThriftyDAgger: Budget-Aware Novelty and Risk Gating for Interactive Imitation Learning},
  author={Ryan Hoque and Ashwin Balakrishna and Ellen R. Novoseller and Albert Wilcox and Daniel S. Brown and Ken Goldberg},
  booktitle={Conference on Robot Learning},
  year={2021},
  url={https://api.semanticscholar.org/CorpusID:237371142}
}

@article{Singi2023DecisionMF,
  title={Decision Making for Human-in-the-loop Robotic Agents via Uncertainty-Aware Reinforcement Learning},
  author={Siddharth Singi and Zhanpeng He and Alvin Pan and Sandip Patel and Gunnar A. Sigurdsson and Robinson Piramuthu and Shuran Song and Matei T. Ciocarlie},
  journal={2024 IEEE International Conference on Robotics and Automation (ICRA)},
  year={2023},
  pages={7939-7945},
  url={https://api.semanticscholar.org/CorpusID:257496009}
}

@article{Ross2012LearningMR,
  title={Learning monocular reactive UAV control in cluttered natural environments},
  author={St{\'e}phane Ross and Narek Melik-Barkhudarov and Kumar Shaurya Shankar and Andreas Wendel and Debadeepta Dey and J. Andrew Bagnell and Martial Hebert},
  journal={2013 IEEE International Conference on Robotics and Automation},
  year={2012},
  pages={1765-1772},
  url={https://api.semanticscholar.org/CorpusID:479635}
}

@article{chi2025diffusion,
  title={Diffusion policy: Visuomotor policy learning via action diffusion},
  author={Chi, Cheng and Xu, Zhenjia and Feng, Siyuan and Cousineau, Eric and Du, Yilun and Burchfiel, Benjamin and Tedrake, Russ and Song, Shuran},
  journal={The International Journal of Robotics Research},
  volume={44},
  number={10-11},
  pages={1684--1704},
  year={2025},
  publisher={Sage Publications Sage UK: London, England}
}

@inproceedings{qureshi2025splatsim,
  title={Splatsim: Zero-shot sim2real transfer of rgb manipulation policies using gaussian splatting},
  author={Qureshi, M Nomaan and Garg, Sparsh and Yandun, Francisco and Held, David and Kantor, George and Silwal, Abhisesh},
  booktitle={2025 IEEE International Conference on Robotics and Automation (ICRA)},
  pages={6502--6509},
  year={2025},
  organization={IEEE}
}

@article{kerbl20233d,
  title={3d gaussian splatting for real-time radiance field rendering.},
  author={Kerbl, Bernhard and Kopanas, Georgios and Leimk{\"u}hler, Thomas and Drettakis, George and others},
  journal={ACM Trans. Graph.},
  volume={42},
  number={4},
  pages={139--1},
  year={2023}
}

@misc{coumans2016pybullet,
  title={Pybullet, a python module for physics simulation for games, robotics and machine learning},
  author={Coumans, Erwin and Bai, Yunfei},
  year={2016}
}

@article{meng2021sdedit,
  title={Sdedit: Guided image synthesis and editing with stochastic differential equations},
  author={Meng, Chenlin and He, Yutong and Song, Yang and Song, Jiaming and Wu, Jiajun and Zhu, Jun-Yan and Ermon, Stefano},
  journal={arXiv preprint arXiv:2108.01073},
  year={2021}
}

@article{zhou2026saps,
  title={SAPS: Shared Autonomy for Policy Steering by Blending Teleoperation with a Pretrained VLA},
  author={Zhou, Crystal and Yang, Jehan and Weber, Douglas J and Erickson, Zackory},
  journal={arXiv preprint arXiv:2606.15568},
  year={2026}
}

@article{jonnavittula2024sari,
  title={SARI: Shared autonomy across repeated interaction},
  author={Jonnavittula, Ananth and Mehta, Shaunak A and Losey, Dylan P},
  journal={ACM transactions on human-robot interaction},
  volume={13},
  number={2},
  pages={1--36},
  year={2024},
  publisher={ACM New York, NY}
}

@article{ho2020denoising,
  title={Denoising diffusion probabilistic models},
  author={Ho, Jonathan and Jain, Ajay and Abbeel, Pieter},
  journal={Advances in neural information processing systems},
  volume={33},
  pages={6840--6851},
  year={2020}
}

@article{pham2018new,
  title={A new approach to time-optimal path parameterization based on reachability analysis},
  author={Pham, Hung and Pham, Quang-Cuong},
  journal={IEEE Transactions on Robotics},
  volume={34},
  number={3},
  pages={645--659},
  year={2018},
  publisher={IEEE}
}

@inproceedings{lee2025diff,
  title={Diff-dagger: Uncertainty estimation with diffusion policy for robotic manipulation},
  author={Lee, Sung-Wook and Kang, Xuhui and Kuo, Yen-Ling},
  booktitle={2025 IEEE International Conference on Robotics and Automation (ICRA)},
  pages={4845--4852},
  year={2025},
  organization={IEEE}
}

\clearpage











\addtolength{\textheight}{-12cm}   








\end{document}